\documentclass[sigconf,natbib=false]{acmart}

\AtBeginDocument{%
  }

\setcopyright{acmcopyright}
\copyrightyear{2018}
\acmYear{2018}
\acmDOI{XXXXXXX.XXXXXXX}

\RequirePackage[
  datamodel=acmdatamodel,
  style=acmnumeric,
  ]{biblatex}

\usepackage{multirow}

\begin{document}

\title{MMNet: Multi-Mask Network for Referring Image Segmentation}

\author{Yichen Yan}
\email{yanyichen2021@ia.ac.cn}
\affiliation{
  \institution{The Laboratory of Cognition and Decision Intelligence for Complex Systems, Institute of Automation, Chinese Academy of Sciences} 
  \city{Beijing}
  \country{China}
}

\author{Xingjian He}
\email{xingjian.he@nlpr.ia.ac.cn}
\affiliation{
  \institution{The Laboratory of Cognition and Decision Intelligence for Complex Systems, Institute of Automation, Chinese Academy of Sciences} 
  \city{Beijing}
  \country{China}
}

\author{Wenxuan Wang}
\email{s20200579@xs.ustb.edu.cn}
\affiliation{
  \institution{School of Automation and Electrical Engineering, University of Science and Technology Beijing} 
  \city{Beijing}
  \country{China}
}

\author{Jing Liu}
\email{jliu@nlpr.ia.ac.cn}
\authornote{corresponding author}
\affiliation{
  \institution{$^{1}$The Laboratory of Cognition and Decision Intelligence for Complex Systems, Institute of Automation, Chinese Academy of Sciences, $^{2}$School of Artificial Intelligence, University of Chinese Academy of Sciences
} 
  \city{Beijing}
  \country{China}
}

\renewcommand\footnotetextcopyrightpermission[1]{}
\settopmatter{printacmref=false} 
\begin{abstract}
Referring image segmentation aims to segment an object referred to by natural language expression from an image. However, this task is challenging due to the distinct data properties between text and image, and the randomness introduced by diverse objects and unrestricted language expression. Most of the previous work have only focused on improving cross-modal feature fusion while not fully addressing the inherent randomness caused by diverse objects and unrestricted language. We propose Multi-Mask Network for referring image segmentation (MMNet), which leverages the Contrastive Language-Image Pretraining (CLIP) to extract both fine-grained and global visual features. To address the randomness, we first combine image and language and then employ an attention mechanism to generate multiple queries that represent different aspects of the language expression. We then utilize these queries to produce a series of corresponding segmentation masks, assigning a score to each mask that reflects its importance. The final result is obtained through the weighted sum of all masks, which greatly reduces the randomness of the language expression. Our proposed framework demonstrates superior performance compared to state-of-the-art approaches on the two most commonly used datasets, RefCOCO, RefCOCO+ and G-Ref, without the need for any post-processing. This further validates the efficacy of our proposed framework.

\end{abstract}


\maketitle

\begin{figure}[t]
  \centering
  \includegraphics[width=\linewidth]{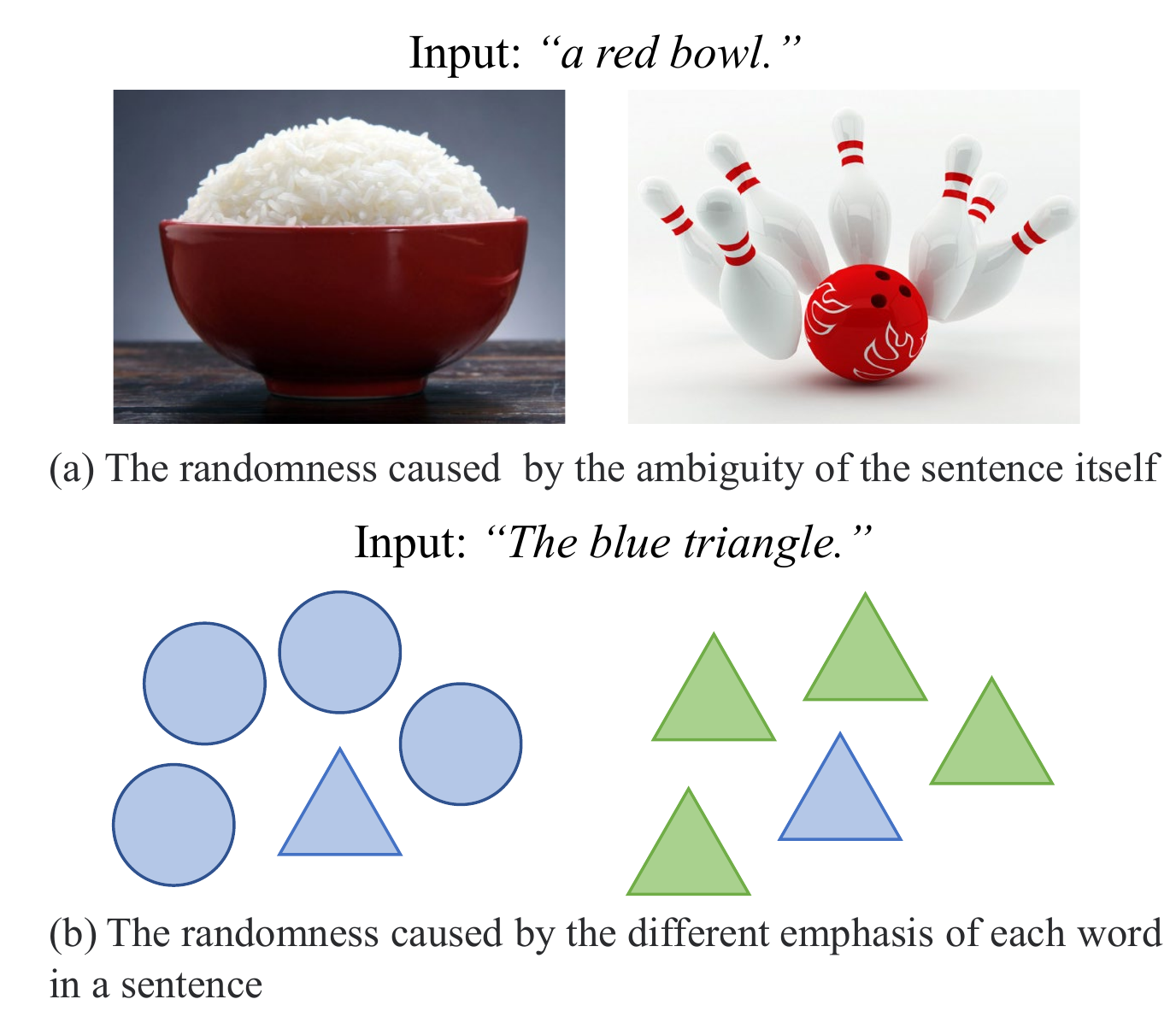}
  \caption{The randomness resulting from the diverse objects/images and unrestricted language expressions}
\end{figure}

\section{Introduction}
Referring image segmentation\cite{hu2016segmentation,liu2017recurrent,li2018referring}, which aims to segment an object referred to by natural language expression from an image, is a fundamental vision-language task. This task has numerous potential applications, including interactive image editing and human-object interaction\cite{wang2019reinforced}. It involves finding a particular region based on the input language expression. This presents two important challenges with this task. The first challenge stems from the distinct data properties between text and image, making it difficult for a network to effectively align text and pixel-level features\cite{wang2022cris}. The second challenge arises from the diverse objects/images and the unconstrained language, leading to a high level of randomness\cite{ding2021vision}. 
\begin{figure*}[t]
  \centering
  \includegraphics[width=\linewidth]{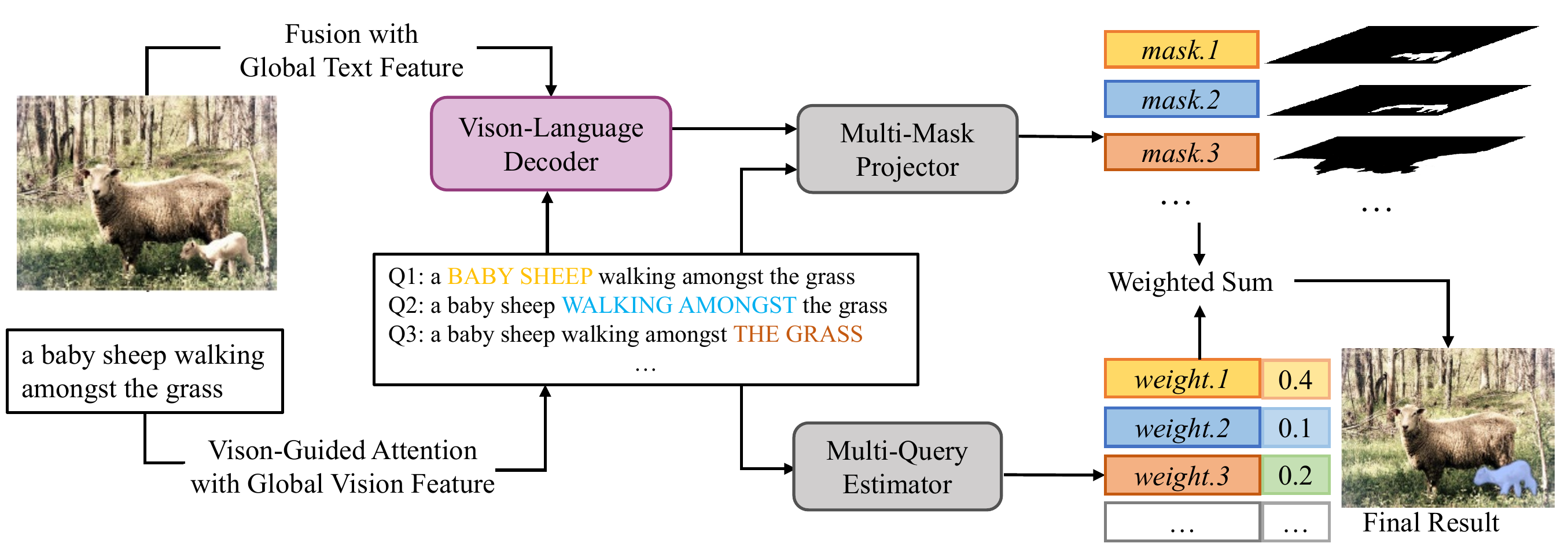}
  \caption{We generated multiple queries and use these queries to obtain corresponding segmentation mask. The final result are obtained by the weighted-sum of these masks}

\end{figure*}

Early works have primarily focused on how to fuse image and language features, with a common approach being the use of concatenation-and-convolution methods to produce the final segmentation result\cite{hu2016segmentation,li2018referring,margffoy2018dynamic}. With the widespread use of attention mechanism several methods have adopted language-vision attention mechanisms to learn cross-modal features more effectively. Recent studies have shown that large-scale pre-trained models can learn high-quality visual representations from natural language supervision, providing a promising alternative to vision-language tasks. Inspired by this, some methods have started using pre-trained models to align the features. Previous work on this task has primarily focused on improving cross-modal feature alignment, but has not fully addressed the inherent randomness resulting from the diverse objects/images and unrestricted language expressions.

The randomness can be explained in two ways. The first is caused by the ambiguity of the sentence itself, as illustrated in Figure 1(a), "bowl" can be a container for food or a kind of ball, We have ambiguity just by looking at the input sentences. Only by combining the image can we know the specific meaning of the input sentence. The second is caused by the different emphasis of each word in a sentence as illustrated in Figure 1(b), for the picture on the left side , if we want to find the object described in the input we should focus more on the shape of the object, while for the picture on the right we should obviously focus more on the color. So the words "blue" and "triangle" are emphasized differently in different images. We cannot address the randomness above just through language expression, so we must combine pictures to give a correct explanation. However, previous methods have often relied on feature extraction from linguistic expressions and images separately, followed by several fusion techniques to obtain multimodal features. These features are then directly sent to the decoder for final result generation through convolution operation which does not possess a deep enough understanding of language to effectively address the randomness inherent in language expressions.

In this paper, we explore addressing the randomness caused by the diverse objects/images and the unconstrained language by generating a series of segmentation masks and eventually combine them to obtain the final result. As illustrated in Figure 2, We generated multiple queries base one language expression. Different from VLT~\cite{ding2021vision}, which generates multiple queries but ultimately uses them to obtain the final mask directly, we generate a corresponding mask for each generated query. The final result is obtained by integrating all the masks which have the same number as queries. These Multi-Mask method can further reduce the impact of randomness. Moreover, we utilize the powerful knowledge of the CLIP\cite{radford2021learning} model to leverage its powerful vision-language knowledge.In summary, our main contributions are listed as follows:
\begin{itemize}
    \item We propose a Multi-Mask Network(MMNet) to produce multiple segmentation and finally use these masks obtain the final result to address the randomness introduced by diverse objects and unrestricted language expression.
    \item We take fully advantage of the CLIP model. Both fine-grained and global visual information of the CLIP are used to improve the performance of our method.  
    \item We test our method on three challenging benchmarks, and we achieve new state-of-the-art results on two of the more difficult datasets RefCOCO+ and G-Ref.
\end{itemize}
\section{Related Work}
\textbf{Referring Image Segmentation.} Referring image segmentation aims at segmenting a specific region in an image by comprehending a given natural language expression\cite{hu2016segmentation}. This task is fundamental and challenging, requiring a deep understanding of both vision and language, and has the potential to be applied in a wide range of domains, such as interactive image editing and human-object interaction. Early works\cite{li2018referring,liu2017recurrent} first extracted visual and textual features by CNN and LSTM, respectively and then and directly concatenated visual and textual features to obtain final segmentation results. Multi-task collaborative network\cite{luo2020multi} achieves a joint learning of referring expression comprehension and segmentation. As the attention mechanism arouses more and more interests, a series of works are proposed to adopt the attention mechanism. For example, BRINet\cite{hu2020bi} applies vision-guided linguistic attention is used to learn the adaptive linguistic context corresponding to each visual region. LAVT\cite{yang2022lavt} conduct multi-modal fusion at intermediate levels of the transformer-based network. VLT\cite{ding2021vision} employs transformer to build a network with an encoder-decoder attention mechanism for enhancing the global context information. CRIS\cite{wang2022cris} employs CLIP\cite{radford2021learning} pretrained on 400M image text pairs and transfers CLIP from text-to-image matching to text-to-pixel matching. However, most of previous work focus on how to improve cross-modal feature fusion effectively, without fully addressing the inherent randomness caused by the diverse objects/images and unrestricted language expressions.

\textbf{Vision-Language Pretraining.} Vision-Language Pretraining models are a type of deep learning models that aim to learn joint representations of both visual and textual information. These models have been shown to be effective in a wide range of natural language processing (NLP) and computer vision (CV) tasks. In recent years, Vision-Language Pretraining has become an important research direction in the field of deep learning\cite{leiless,lu2019vilbert,su2019vl,devlin2018bert,li1908universal}.As a milestone, CLIP\cite{radford2021learning} employs a contrastive learning strategy on a huge mount of image-text pairs, and shows impressive transferable ability over 30 classification datasets. Motivated by this work, a number of follow-ups\cite{fang2021clip2video,luo2022clip4clip,patashnik2021styleclip,tang2021clip4caption,wang2022clip} have been proposed to transfer the knowledge of CLIP models to downstream tasks and achieved promising results. CRIS\cite{wang2022cris} aims to transfer image-level visual concepts to referring image segmentation to leverage multi-modal corresponding information. However, the approach only focuses on fine-grained visual representations and neglects the importance of global visual information, which is a critical aspect of CLIP. Because the global visual information, combined with global textual information, is used to calculate the contrastive score, which is the core of CLIP. Our approach, on the other hand, also employs CLIP to address referring image segmentation, but with a focus on both fine-grained and global visual information. By incorporating both types of information, we aim to improve the accuracy and effectiveness of the referring image segmentation task.  
\begin{figure*}[t]
  \centering
  \includegraphics[width=\linewidth]{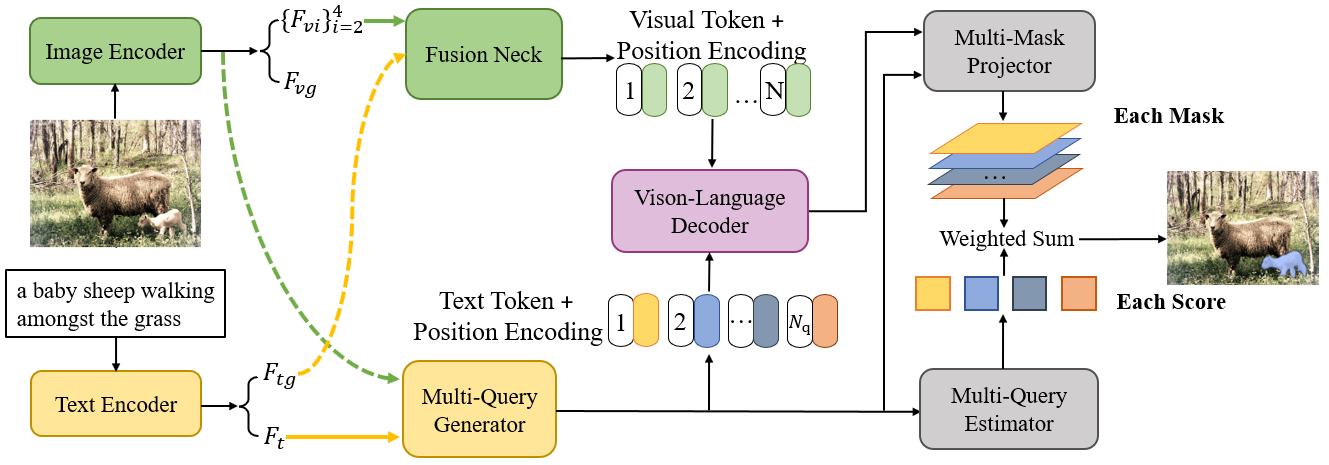}
  \caption{We generated multiple queries and use these queries to obtain corresponding segmentation mask. The final result are obtained by the weighted-sum of these masks}
\end{figure*}

\section{Methodology}

As illustrated in Figure 3, our proposed framework facilitates knowledge transfer to generate multiple queries and their corresponding masks to obtain the final prediction. Firstly, the framework takes an image and a language expression as input. We employ ResNet/ViT and a Transformer to extract image and text features, respectively. Both modalities' features, including global features, are extracted. The global text feature and the vision features are further fused to obtain simple multi-modal features. To generate multiple queries, we utilize the global vision feature, the patch features, and the text features in Multi-Query Generator.

Secondly, the generated multiple queries and multi-modal features are input into the vision-language decoder. The decoder's output, along with the generated queries, are fed into the Multi-Mask Projector to produce multiple masks, with each query's help. Meanwhile, the Multi-Query Estimator uses the generated queries to determine the weights of each mask produced by the Multi-Mask Projector. Finally, we use those masks and their corresponding weights to calculate the weighted sum to obtain the final prediction.

\subsection{Image and Text Feature Extraction}

\textbf{Text Encoder}. For a given language expression $\emph{T} \in \mathbb{R}^{\emph{L}}$, we utilize a Transformer to obtain text features $\emph{$F_t$} \in \mathbb{R}^{\emph{L}\,\times\,\emph{C}}$. We follow CLIP's approach and use byte pair encoding (BPE) to begin the text sequence with the \verb|[SOS]| token and end it with the \texttt{[EOS]} token. Additionally, like CRIS, we use the highest layer's activations of the Transformer at the \texttt{[EOS]} token as the global feature for the entire language expression. This feature is linearly transformed and denoted as $F_{tg} \in \mathbb{R}^{C^{\prime}}$. Here, \emph{C} and \emph{$C^{\prime}$} represent the feature dimension, while L is the length of the language expression.

\textbf{Image Encoder}. For a given image $I \in \mathbb{R}^{H \times W \times 3}$, we not only extract its visual features but also its global visual representation, unlike CRIS. As an example, we use the ResNet encoder.  In architecture of the CLIP image encoder, they take the ResNet encoder, there are 4 stages in total and we denote the feature as $\left\{\mathbf{x}_{i}\right\}_{i=1}^{4}$. Unlike the original ResNet, CLIP adds an attention pooling layer. Specifically, CLIP applies global average pooling to $\mathbf{x}_{4} \in \mathbb{R}^{H_{4} \times W_{4} \times C}$ to obtain a global feature, denoted as $\overline{\mathbf{x}}_4 \in \mathbb{R}^{C}$. Here, $H_4$, $W_4$, and $C$ are the height, width, and number of channels of $\mathbf{x}_4$. Then, CLIP concatenates the features $\left[\overline{\mathbf{x}}_4, \mathbf{x}_4\right]$ and feeds them into a multi-head self-attention layer.
\begin{equation}
[\overline{\mathbf{z}}, \mathbf{z}]=MHSA\left(\left[\overline{\mathbf{x}}_4, \mathbf{x}_4\right]\right).
\end{equation}
In CLIP model, the final output of the image encoder is the global visual feature $\overline{\mathbf{z}}$, which is used to calculate the contrastive score with the global textual feature from the Text Encoder. Other outputs $\mathbf{z}$, are typically disregarded. In contrast, the CRIS model utilizes the other output $\mathbf{z}$ as a feature map due to its adequate spatial information, while the global visual feature $\overline{\mathbf{z}}$ is discarded.

In our proposed model, we leverage the multiple visual features $\mathbf{x}_{2} \in \mathbb{R}^{H_{2} \times W_{2} \times C}$ and $\mathbf{x}_{3} \in \mathbb{R}^{H_{3} \times W_{3} \times C}$ from the second and third stages of the ResNet, respectively, similar to CRIS. We transform them into $F_{v 2} \in \mathbb{R}^{H_{2} \times W_{2} \times C_2}$ and $F_{v 3} \in \mathbb{R}^{H_{3} \times W_{3} \times C_3}$ with two learnable matrices, respectively. However, in the fourth stage, we differ from both CLIP and CRIS. We not only use the visual feature $\mathbf{z}$ to extract sufficient spatial information, but also incorporate the global visual feature $\overline{\mathbf{z}}$ to capture the global information of the image. To accomplish this, we employ two  learnable projection matrices to transform $\mathbf{z}$ and $\overline{\mathbf{z}}$ into $F_{v 4} \in \mathbb{R}^{H_{4} \times W_{4} \times C_4}$ and $F_{v g} \in \mathbb{R}^{C_4}$. It is also noted that for architectures like ViT\cite{dosovitskiy2020image}, z can be obtained similarly by excluding the class token of outputs.

\textbf{Fusion Neck}. In the Fusion Neck, we perform a straightforward fusion of multiple features, including $F_{v 2}$, $F_{v 3}$, $F_{v 4}$, and the global textual feature $F_{t g}$, to generate a visual feature that incorporates global textual information. Initially, we fuse $F_{v 4}$ and $F_{t g}$ to obtain $F_{m 4} \in \mathbb{R}^{H_{3} \times W_{3} \times C}$ using the following equation:
\begin{equation}
F_{m 4}=U p\left(\sigma\left(F_{v 4} W_{v 4}\right) \cdot \sigma\left(F_{tg} W_{tg}\right)\right),
\end{equation}
In this process, $Up(\cdot)$ denotes 2 $\times$ upsampling function, and $\cdot$ denotes element-wise multiplication. We first transform the visual and textual representations into the same feature dimension using two learnable matrices, $W_{v 4}$ and $W_{tg}$, and then We apply ReLU activation function which is denoted as $\sigma$ to generates $F_{v 4}$.
Subsequently, we obtain the multi-modal features $F_{m 3}$ and $F_{m 2}$ using the following procedures:
\begin{equation}
\begin{aligned}
& F_{m_3}=\left[\sigma\left(F_{m_4} W_{m_4}\right), \sigma\left(F_{v_3} W_{v_3}\right)\right], \\
& F_{m_2}=\left[\sigma\left(F_{m_3} W_{m_3}\right), \sigma\left(F_{v_2}^{\prime} W_{v_2}\right)\right], F_{v_2}^{\prime}=Avg\left(F_{v_2}\right),
\end{aligned}
\end{equation}
Where $Avg(\cdot)$ denotes a kernel size of 2 $\times$ 2 average pooling operation
with 2 strides, $[,]$ denotes the concatenation operation.Subsequently, we concatenate the three multi-modal features ($F_{m 4}, F_{m 3}, F_{m 2}$) and use a $1 \times 1$ convolution layer to aggregate them:
\begin{equation}
F_m= Conv\left(\left[F_{m_2}, F_{m_3}, F_{m_4}\right]\right),
\end{equation}
Where $F_{m} \in \mathbb{R}^{H_{3} \times W_{3} \times C}$. Then, we obtain the 2D spatial coordinate feature $F_{coord} \in \mathbb{R}^{H_{3} \times W_{3} \times 2}$ and concatenate it with $F_{m}$ and flatten the result to obtain the fused visual features with global textual information which is denoted as $F_{vt} \in \mathbb{R}^{H_{3}W_{3} \times C}$.
\begin{equation}
F_{vt}=Flatten\left(Conv\left(\left[F_m, F_{coord}\right]\right)\right).
\end{equation}
Here, $Flatten(\cdot)$ denotes flatten operation and we obtain the $F_{vt} \in \mathbb{R}^{N \times C}$, $N = H_{3} \times W_{3} = \frac{H}{16} \times \frac{W}{16}$, which will be utilized in the following process. As for ViT\cite{dosovitskiy2020image}, we will directly extract its class token as a global visual feature and then use three convolution to obtain the three features which have the same dimension as $F_{v2}$, $F_{v3}$ and $F_{v4}$. After that, the operation is the same as ResNet.

\subsection{Multi-Query Generator}Similar to the VLT\cite{ding2021vision}, the Multi-Query Generator is designed to generate a series of queries that represent different interpretations of the image to However, unlike VLT, which only utilizes visual and textual features to directly generate queries, our approach incorporates both detailed visual features and holistic global visual features to guide the query generation process.

As shown in Figure 3, the Multi-Query Generator takes multiple stage visual features $\left\{F_{vi}\right\}_{i=2}^4$, the global visual feature $F_{vg}$, and textual features ${F_t}$ as input and outputs a series of queries. To generate multiple queries, we first need to obtain dense visual features and fused textual features.
  
 \textbf{Dense Visual Features}. In according to obtain the dense visual Features, we use the multiple stage visual features $\left\{F_{vi}\right\}_{i=2}^4$. The operations we take are very similar to those in Fusion Neck of the Image and Text Feature Extraction, but the difference is that we do not use element-wise multiplication with global textual feature in the first step of the Fusion Neck. And we obtain the dense visual features by following process:
\begin{equation}
\begin{aligned}
& F_{m_4}^{\prime}=U p\left(\sigma\left(F_{v 4} W_{v 4}^{\prime}\right)\right), \\
& F_{m_3}^{\prime}=\left[\sigma\left(F_{m_4}^{\prime} W_{m_4}^{\prime}\right), \sigma\left(F_{v_3} W_{v_3}^{\prime}\right)\right], \\
& F_{m_2}^{\prime}=\left[\sigma\left(F_{m_3}^{\prime} W_{m_3}^{\prime}\right), \sigma\left(F_{v_2}^{\prime\prime} W_{v_2}^{\prime}\right)\right], F_{v_2}^{\prime\prime}=Avg\left(F_{v_2}\right),\\
& F_m^{\prime}= Conv\left(\left[F_{m_2}^{\prime}, F_{m_3}^{\prime}, F_{m_4}^{\prime}\right]\right), F_{v}^{\prime}=Conv\left(\left[F_m^{\prime}, F_{coord}\right]\right),
\end{aligned}
\end{equation}
Just like the operation in Fusion Neck, $Up(\cdot)$ denotes 2 $\times$ upsampling, $Avg(\cdot)$ denotes a kernel size of 2 $\times$ 2 average pooling operation
with 2 strides, $[,]$ denotes the concatenation operation. Here, $F_{v d} \in \mathbb{R}^{ H_{3} \times W_{3} \times C}$, just like the  $F_{v t}$ without flattening, but the difference is that $F_{v d}$ does not incorporate the global textual information. 

After obtaining the dense visual features $F_{v d}$, we apply three convolution layers to reduce the feature channel dimension size to the desired number of queries $N_q$. This results in $N_q$ feature maps, which are flattened in the spatial domain. Specifically, each feature map is flattened to a one-dimensional vector of length $H_{3} \times W_{3}$, where $H_{3}$ and $W_{3}$ are the height and width of the dense visual feature maps, respectively. This results in a matrix of size $N_q \times H_{3}W_{3}$, which contains detailed visual information. And the specific process is follow:
\begin{equation}
F_{v d}=flatten\left(Conv\left(F_{v}^{\prime}\right)\right)^T,
\end{equation}

\textbf{Fused textual features}. Unlike VLT, which uses raw textual features obtained by the Text Encoder, we use fused textual features which incorporate the global visual feature. We fuse the textual features and the global visual feature by following equation:
\begin{equation}
F_{t v}=\sigma\left(F_t W_{t}\right)\cdot\sigma\left(F_{vg} W_{vg}\right),
\end{equation}
Here, $F_{t v}\in \mathbb{R}^{\emph{L} \times \emph{C}}$, $W_t$ and $W_{vg}$ are two learnable matrices.

\textbf{Multi-Query Generation}. For referring image segmentation, the importance of different words in the same language expression is obviously different. Some previous works address this issue by measuring the importance of each word and give each word a weight by the language self-attention. But what they neglect is that the importance of different words in the same language expression can vary depending on the specific image being referred to. About this, the VLT\cite{ding2021vision} makes a detailed explanation. Therefore, we need to combine the language expression with the visual information to generate a set of queries that are specific to the given image. In our approach, we use the dense visual features and the fused textual features to generate multiple queries, each corresponding to a different interpretation of the image. We do this by computing attention weights between the dense visual features and the fused textual features for each query, which helps to determine the relevance of different words in the language expression for each query. 

In order to derive attention weights for fused textual features $F_{tv}$, we incorporate the dense vision features $F_{vd}$, as illustrated in Fig. 5. Following the approach of VLT, we begin by applying linear projection to $F_{vd}$ and $F_{tv}$. Then, for the $n$-th query ($n = 1,2,...,N_q$), we take the $n$-th dense visual feature vector $f_{vdn} \in \mathbb{R}^{1 \times \left(H_3W_3\right)}$ , along with the fused textual features of all words. Specifically, we use $f_{tvi} \in \mathbb{R}^{1 \times C}$ to denote the feature of the $i$-th word ($i = 1,2,...,L$). The attention weight for the $i$-th word with respect to the $n$-th query is computed as the product of projected $f_{vdn}$ and $f_{tvi}$:
\begin{equation}
a_{n i}=\sigma\left(f_{vdn} W_{vd}\right) \sigma\left(f_{tvi} W_{a}\right)^T,
\end{equation}
In the equation, $a_{ni}$ represents a scalar that indicates the importance of the $i$-th word in the $n$-th query, where $W_{vd}$ and $W_{a}$ are learnable matrices. To normalize the attention weights across all words for each query, we apply the Softmax function. The resulting values of $a_{ni}$, after being processed by Softmax, comprise the attention map $A \in \mathbb{R}^{N_q \times L}$. For the $n$-th query, we extract $A_n \in \mathbb{R}_{1 \times L}$ ($n = 1,2,...,N_q$) from A, which represents the emphasis of the words on the $n$-th query. And $A_n$ are use to generate the new queries as following equation:
\begin{equation}
    F_{qn} = A_n\sigma\left(F_{t v} W_{t v}\right). 
\end{equation} 
The matrix $W_{tv}$ is a learnable parameter. The feature vector $F_{qn} \in \mathbb{R}^{1 \times C}$ is guided by both dense visual information and global visual information, additionally, each new query is a projected weighted sum of the features of different words in the language expression. This enables the query to retain its properties as a language feature and allows it to be used to query the image, so it can serve as a single query vector for the Vision-Language Decoder. The set of all queries comprises the new language matrix $F_q \in \mathbb{R}^{N_q \times C}$, which is called generated query matrwill be input to the Vision-Language Decoder.

\subsection{Vision-Language Decoder}
We employ a Vision-Language Decoder to facilitate the transfer of fine-grained semantic information from textual features to visual features in an adaptive manner. As illustrated in Figure 3, the decoder takes query vectors $F_q$ and fused visual features $F_{vt}$ as input. To incorporate positional information, we add $F_{vt}$\cite{carion2020end} and $F_q$\cite{vaswani2017attention} with sine spatial positional encodings. The decoder architecture follows the standard transformer\cite{vaswani2017attention} design, where each layer consists of a multi-head self-attention layer, a multi-head cross-attention layer, and a feed-forward network. In each decoder layer, the multi-head self-attention layer is applied to $F_{vt}$ to capture global contextual information:
\begin{equation}
    F_{vt}^{\prime} = MHSA\left(LN\left(F_{vt}\right)\right) + F_{v t}^{\prime}.
\end{equation}
The resulting evolved visual feature is denoted as $F_{vt}^{\prime}$, where $MHSA(\cdot)$ and $LN(\cdot)$ represent the multi-head self-attention layer and Layer Normalization\cite{ba2016layer}, respectively. The multi-head self-attention mechanism consists of three linear layers that map $F_{vt}$ to intermediate representations, including queries $Q \in \mathbb{R}^{N \times d_q}$, keys $K \in \mathbb{R}^{N \times d_k}$, and values $V \in \mathbb{R}^{N \times d_v}$. The multi-head self-attention calculation is then expressed as:
\begin{equation}
    MHSA\left(Q,K,V\right) = softmax\left(\frac{QK^T}{\sqrt{d_k}}\right).
\end{equation}
Subsequently, we use a multi-head cross-attention layer to propagate fine-grained semantic information into the evolved visual features. Here, $Q$ is obtained by a linear projection of $F_{vt}^{\prime}$, while $K$ and $V$ are both derived by two separate linear projections of $F_{q}$. To obtain the multi-modal feature $F_s$, the output query $Q$ is processed through an MLP block comprising two layers with Layer Normalization and residual connections:

\begin{equation}
\begin{aligned}
& F_s^{\prime}=M H C A\left(L N\left(F_{v t}^{\prime}\right), F_{q}\right)+F_{v t}^{\prime}, \\
& F_s=M L P\left(L N\left(F_s^{\prime}\right)\right)+F_s^{\prime}.
\end{aligned}
\end{equation}
Here, $MHCA(\cdot)$ denotes the multi-head cross-attention layer, and $F_s^{\prime}$ represents the intermediate features. The evolved multi-modal feature $F_s$ is utilized to generate the final segmentation mask.
\begin{table*}[thbp]
    \setlength{\belowcaptionskip}{1.0pt}
    \begin{center}
    \caption{\textbf{Comparisons with the state-of-the-art approaches on three benchmarks.}
    We report the results of our method with various visual backbones.
    ``$\star$'' denotes the post-processing of DenseCRF \cite{krahenbuhl2011efficient}.
    ``$\dag$'' denotes the Swin Transformer\cite{liu2021swin} pre-trained on ImageNet-22K\cite{deng2009imagenet}.
    ``-'' represents that the result is not provided.
    IoU is utilized as the metric.}
    \setlength{\tabcolsep}{2.8mm}{
    \begin{tabular}{l|c|ccc|ccc|cc}
        \toprule[1.2pt]
        \multirow{2}{*}{Method} & \multirow{2}{*}{Backbone} & \multicolumn{3}{c|}{RefCOCO} & \multicolumn{3}{c|}{RefCOCO+} & \multicolumn{2}{c}{G-Ref} \\
        \cline{3-10}
        ~ & ~ & val & test A & test B & val & test A & test B & val & test \\
        \midrule[1.2pt]

        RRN$^\star$ \cite{li2018referring}     & ResNet-101 & 55.33 & 57.26 & 53.95 & 39.75 & 42.15 & 36.11 & - & - \\
        MAttNet \cite{yu2018mattnet}           & ResNet-101 & 56.51 & 62.37 & 51.70 & 46.67 & 52.39 & 40.08 & 47.64 & 48.61 \\
        MCN \cite{luo2020multi}                & DarkNet-53 & 62.44 & 64.20 & 59.71 & 50.62 & 54.99 & 44.69 & 49.22 & 49.40 \\
        CGAN \cite{luo2020cascade}             & DarkNet-53 & 64.86 & 68.04 & 62.07 & 51.03 & 55.51 & 44.06 & 51.01 & 51.69 \\
        EFNet \cite{feng2021encoder}           & ResNet-101 & 62.76 & 65.69 & 59.67 & 51.50 & 55.24 & 43.01 & - & - \\
        LTS \cite{jing2021locate}              & DarkNet-53 & 65.43 & 67.76 & 63.08 & 54.21 & 58.32 & 48.02 & 54.40 & 54.25 \\
        VLT \cite{ding2021vision}                 & DarkNet-53 & 65.65 & 68.29 & 62.73 & 55.50 & 59.20 & 49.36 & 52.99 & 56.65 \\
        ReSTR\cite{kim2022restr}                    &  ViT-B &  67.22 & 69.30 & 64.45 & 55.78 & 60.44 &48.27 & 54.48 & - \\
        SeqTR\cite{li2021referring}                 & ResNet-101  & 70.56 & 73.49 & 66.57 & 61.08 & 64.69 & 52.73 & 58.73 & 58.51 \\
        CRIS\cite{wang2022cris}                           & ResNet-101 & 70.47 & 73.18 & 66.10 & 62.27 & 68.08 & 53.68 & 59.87 & 60.36 \\
        LAVT$^\dag$\cite{yang2022lavt}                           & Swin-B  & \underline{72.73} & \underline{ 75.82} & \underline{68.79} & 62.14 & 68.38 & 55.10 & 61.24 & 62.09 \\
        \midrule
        MMNet(Ours)                     &ViT-B &71.74 & 74.38 & 67.24 & \underline{64.05} & \underline{68.41} & \underline{56.82} & \underline{61.87} & \underline{62.21}\\
        MMNet(Ours)                    &ViT-L & \textbf{75.01}& \textbf{77.81} & \textbf{71.59} & \textbf{68.44} & \textbf{72.81} & \textbf{59.86} & \textbf{66.52} & \textbf{67.28}  \\
        \bottomrule[1.2pt]
    \end{tabular}
    \label{tab:sota}}
    \end{center}
    \vspace{-5.0mm}
\end{table*}

\subsection{Mask Decoder}
In contrast to VLT\cite{ding2021vision}, which aggregates the information of all queries to obtain a single mask as the final result, we leverage the information of each query to generate a mask for each one. We then aggregate the resulting $N_q$ masks to obtain the final output. By doing so, we make full use of the information contained in each query, leading to a more nuanced and precise understanding of the input language expression. 

Specifically, We have obtained evolved multi-modal features, which is denoted as $F_s$. Simultaneously, we have generated $N_q$ queries. For each query, we generate a segmentation mask combined with $F_s$, resulting in a total of $N_q$ masks. This process occurs at the Multi-Mask Projector, and each mask represents a specific comprehension of the input language expression. As we previously discussed, both the input image and language expression exhibit a high degree of randomness. Therefore, it is desirable to adaptively select the most appropriate comprehension ways, allowing the network to focus on the most reasonable and suitable ones. Furthermore, given the independence of each query vector in the transformer decoder, but with only one mask output desired, it is necessary to balance the influence of different queries on the final output. Specifically, we feed each query into the Multi-Query Estimator, which evaluates it and assigns a score reflecting the quality of the mask generated by this query. We then use these scores to weight and sum all the masks, resulting in the final mask. 

\textbf{Multi-Mask Projector}. As illustrated in Figure 3, Multi-Mask Projector takes multi-modal feature $F_s$ and query vectors $F_q$ as input. We extract one query $F_{q n}$ from $F_q$, and $F_{q n}$ is used to generate a mask with the help of $F_s$. We use a dynamic convolution operation\cite{chen2020dynamic}, and the parameters of the convolution kernel come from $F_{q n}$. The detailed operations are as follows:
\begin{equation}
\begin{aligned}
& F_p = Up(Conv(Up(F_s))), \\
& F_{p n} = \sigma(W_pF_{q n}),
\end{aligned}
\end{equation}
Here, we use 2 $\times$ unsampling and convolution operation to transform $F_s$ into  $F_p \in \mathbb{R}^{4H_3 \times 4W_3 \times C_p}$, $C_p = \frac{C}{2}$. Then we use a linear layer to transform $F_{q n}$ into $F_{p n} \in \mathbb{R}^{9C_p + 1}$. From the vector $F_{pn}$, we take the first $9C_p$ values as parameters of the $3 \times 3$ convolution kernel whose the number of channel is $C_p$, and we choose the last value of $F_{pn}$ as bias, and then we utilize convolution to obtain a mask generated by the $n$-th query $F_{q n}$, which is denoted as $mask_n \in \mathbb{R}^{4H_3 \times 4W_3 \times 1}$. 

\textbf{Multi-Query Estimator}. As illustrated in Figure 3, the Multi-Query Estimator takes the query vectors $F_q$ as input and outputs $N_q$ scores. Each score shows how much the query $F_{q n}$ fits the context of its prediction, and controls the influence of its response $mask_n$ generated by the itself. The Multi-Query Estimator first applies a multi-head self-attention layer and then employs a linear layer to obtain $N_q$ scalar:
\begin{equation}
    S_q = Softmax(W_s(MHSA(F_q))),
\end{equation}
Here, $S_q \in \mathbb{R}^{N_q \times 1}$. The linear layer uses Softmax as an activation function to control the output range. The final prediction is derived from the weighted sum of the mask obtained by the Multi-Mask Generator and the score obtained by the Multi-Query Estimator:
\begin{equation}
    y = \sum_{n=1}^{N_q} S_{qn}mask_n.
\end{equation}
Here, $S_{qn}$ is $n$-th scalar of the $S_q$, $y$ denotes the final prediction mask. The model is optimized with cross-entropy loss.

\section{Experiments}
\subsection{Implementation Details}
\textbf{Experiment Settings.} We strictly follow previous works\cite{wang2022cris} for experiment settings, including preparing the ResNet-101\cite{he2016deep} and ViT\cite{dosovitskiy2020image} as the image encoder. Input images are resized to $480 \times 480$. Due to the extra \texttt{[SOS]} and \texttt{[EOS]} tokens, and the input sentences are set with a maximum sentence length of 17 for RefCOCO and RefCOCO+, and 22 for G-Ref. Each Transformer block has 8 heads, and the hidden layer size in all heads is set to 512, and the feed-forward hidden dimension is set to 2048. We train the network for 100 epochs using the Adam optimizer with the learning rate lr = 1e-5 and decreases with polynomial decay\cite{chen2014semantic}. We train the model with a batch size of 64 on 8 RTX Titan with 24 GPU VRAM.
\textbf{Metrics.} Following previous works\cite{wang2022cris,ding2021vision,yang2022lavt}, we adopt two metrics to verify the effectiveness: IoU and Precision@$X$.The IoU calculates intersection regions over union regions of the predicted segmentation mask and the ground truth. The Precision@$X$ measures the percentage of test images with an IoU score higher than the threshold $X \in \{0.5, 0.6, 0.7, 0.8, 0.9\}$, which focuses on the location ability of the method.
\subsection{Datasets}
We conduct our method on three standard benchmark datasets, RefCOCO\cite{yu2016modeling}, RefCOCO+\cite{yu2016modeling}, and G-Ref\cite{nagaraja2016modeling}, which are widely used in referring image segmentation task. Images in the three datasets are collected from the MS COCO dataset\cite{lin2014microsoft} and annotated with natural language expressions. RefCOCO and RefCOCO+ are among the largest image datasets for referring segmentation. The RefCOCO dataset contains 142,209 referring language expressions describing 50,000 objects in 19,992 images, while the RefCOCO+ dataset contains 141,564 referring language expressions for 49,856 objects in 19,992 images. The main difference between RefCOCO and RefCOCO+ is that RefCOCO+ only contains appearance expressions, and does not include words that indicate location properties (such as left, top, front) in expressions. G-Ref is another prominent referring segmentation dataset that contains 104,560 referring language expressions for 54,822 objects across 26,711 images. Unlike RefCOCO and RefCOCO+, the language usage in the G-Ref is more casual but complex, and the sentence lengthes of G-Ref are also longer in average. Furthermore, the G-Ref dataset has two partitions: one created by UMD\cite{nagaraja2016modeling} and the other by Google\cite{mao2016generation}. In our paper, we report results on the UMD partition.
\begin{figure}[htbp]
  \centering
  \includegraphics[width=\linewidth]{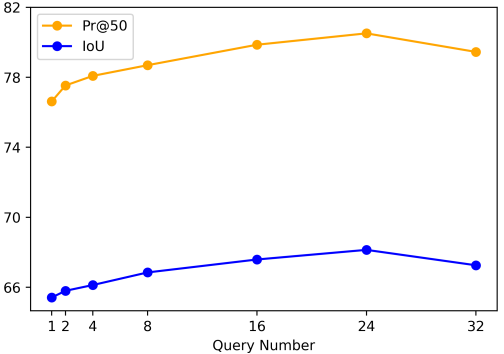}
  \caption{Performance gain by different query number $N_q$}
  \vspace{-5.0mm}
\end{figure}
\begin{table}[htbp]
    \vspace{-0.2cm} 
    \setlength{\belowcaptionskip}{1.0pt}
    \begin{center}
    
    \caption{\textbf{Influence of Query Numbers.}}
    \setlength{\tabcolsep}{1.8mm}{
    \begin{tabular}{c|c|c|c|c|c|c}
        \toprule[1.2pt]
        $N_q$ & IoU & Pr@50 & Pr@60 & Pr@70 & Pr@80 & Pr@90 \\
        \midrule
          32 & 67.26 & 79.45 & 75.58 & 68.71 & 52.65 & 14.91 \\
          24 & \textbf{68.14} & \textbf{80.51} & \textbf{76.89} & \textbf{70.22}  & \textbf{52.99}  & \textbf{15.46} \\
          16 & 67.59 & 79.86 & 75.82 & 69.03 & 52.71 & 14.80 \\
          8  & 66.85 & 78.69 & 75.03 & 68.05 & 52.55 & 14.83 \\
          4  & 66.13 & 78.08 & 74.39 & 68.54 & 52.12 & 14.29 \\
          2  & 65.80 & 77.53 & 74.19 & 68.13 & 51.87 & 14.39 \\
          1  & 65.42 & 76.62 & 72.73 & 66.26 &51.41  & 13.84 \\
        \bottomrule[1.2pt]
    \end{tabular}
    \label{tab:Nq}}
    \end{center}
\end{table}

\subsection{Compare with others}
In Table 1, we evaluate MMNet against the state-of-the-art referring image segmentation methods on the RefCOCO\cite{yu2016modeling}, RefCOCO+\cite{yu2016modeling}, and G-Ref\cite{nagaraja2016modeling} datasets using the IoU metric. We first use the basic visual backbone ViT-Base. Results show that our proposed method outperforms other methods on RefCOCO+ and G-Ref datasets. On the RefCOCO+ dataset, our method achieves higher IoU performance than other methods. Compared to the second-best performing method, LAVT\cite{yang2022lavt}, our MMNet model achieves absolute margins of 1.91, 0.03, and 1.72 scores on the validation, testA, and testB subsets of RefCOCO+, respectively. Our proposed method also outperforms LAVT on the more complex G-Ref dataset with 0.63 and 0.12 absolute score improvement. On RefCOCO datasets, we get a comparable result with other methods. To further validate the potential of our model, we also conduct additional experiments using a more robust visual backbone ViT-Large. Compared with the second-best method LAVT, our method achieves higher performance with absolute margins of
3.13\%, 2.62\%, and 4.07\% on the validation, testA, and
testB subsets of RefCOCO, respectively. Similarly, our method
attains noticeable improvements over the previous state of
the art on RefCOCO+ with wide margins of 10.14\%, 6.48\%,
and 8.64\% on the validation, testA, and testB subsets, respectively. On the G-Ref dataset, our method surpasses the second-best methods on the validation and
test subsets from the UMD partition by absolute margins
of 8.62\% and 8.36\%, respectively.Specifically, our model performs better on datasets with relatively difficult language expressions, RefCOCO+ and G-Ref, demonstrating its ability to understand challenging language expressions from different aspects and effectively deal with their inherent randomness.
\subsection{Ablation Study}
We conduct several ablations to evaluate the effectiveness of the key components in our proposed network. we do the ablation study on a
more difficult dataset, the testA split of RefCOCO+, we use ResNet50 as the vision backbone and the epoch is set to 50.

\textbf{Query Number}. In order to clarify  the influence of the query number $N_q$, we set the $N_q$ to a series of different number. The result are reported at Table 2 and Figure 4. According to the result, multiple queries can improve the performance of our model which is about $4\%$ from 1 query to 24 queries. The result of Pr@50 also shows that the significant performance brought by the multiple queries. This also shows that multiple queries generated by the Multi-Query Generator represent different aspects of information and we can obtain a good multi-modal features. However,  more $N_q$ is not always bring a better result, With the increase of $N_q$, the performance will gradually level off or even decline.
\begin{table}[htbp]
    \begin{center}
    \caption{\textbf{Comparison of whether to use Multi-Mask Projector(MMP) to produce multiple masks}}
    \setlength{\tabcolsep}{1.2mm}{
    \begin{tabular}{c|c|c|c|c|c|c|c}
        \toprule[1.2pt]
        $N_q$ & MMP &IoU & Pr@50 & Pr@60 & Pr@70 & Pr@80 & Pr@90 \\
        \midrule
          \multirow{2}{*}{24} & \checkmark & \textbf{68.14}   &\textbf{80.51}  & \textbf{76.89} & \textbf{70.22} & \textbf{52.99} & \textbf{15.46} \\
            \cline{2-2}
          ~ & & 66.82  &78.59 & 74.90 & 68.78 & 52.28 & 15.18 \\
          \midrule
          \multirow{2}{*}{16} & \checkmark & 67.59 & 79.86 & 75.82 & 69.03 & 52.71 & 14.80 \\
          \cline{2-2}
          ~ & & 66.65 & 78.24 & 74.85 & 67.91 & 50.56 & 13.87 \\
          \midrule
          \multirow{2}{*}{8} & \checkmark & 66.85 & 78.69 & 75.03 & 68.05 & 52.55 & 14.83 \\
          \cline{2-2}
          ~ & & 66.09 & 77.38 & 73.89 & 67.05 & 49.60 & 13.41 \\
        \bottomrule[1.2pt]
    \end{tabular}
    \label{tab:Nq}}
    \end{center}
    \vspace{-2.0mm}
\end{table}

\begin{table}[htbp]
    \begin{center}
    \caption{\textbf{Other ablation results on the RefCOCO+ testA set. MQE denote the Multi-Query Estimator}}
    \setlength{\tabcolsep}{1.4mm}{
    \begin{tabular}{c|c|c|c|c|c|c|c}
        \toprule[1.2pt]
        $f_{vg}$ & MQE & IoU & Pr@50 &Pr@60 & Pr@70 & Pr@80 & Pr@90 \\
        \midrule
        \checkmark & \checkmark &\textbf{68.14}  &\textbf{80.51} & \textbf{76.89} & \textbf{70.22} &\textbf{52.99} &\textbf{15.46} \\
         ~    & \checkmark & 67.32  &79.22 & 75.83 & 69.39 & 53.48 & 14.91\\
         \checkmark & ~ & 66.69  &78.54 & 75.35 & 69.03 & 52.43 & 14.73 \\
         ~ & ~ & 66.17  &77.46 & 74.07 & 67.91 & 51.91 & 14.38\\
        \bottomrule[1.2pt]
    \end{tabular}
    \label{tab:Nq}}
    \end{center}
    \vspace{-5.0mm}
\end{table}
\begin{figure*}[thbp]
  \centering
  \includegraphics[width=\linewidth]{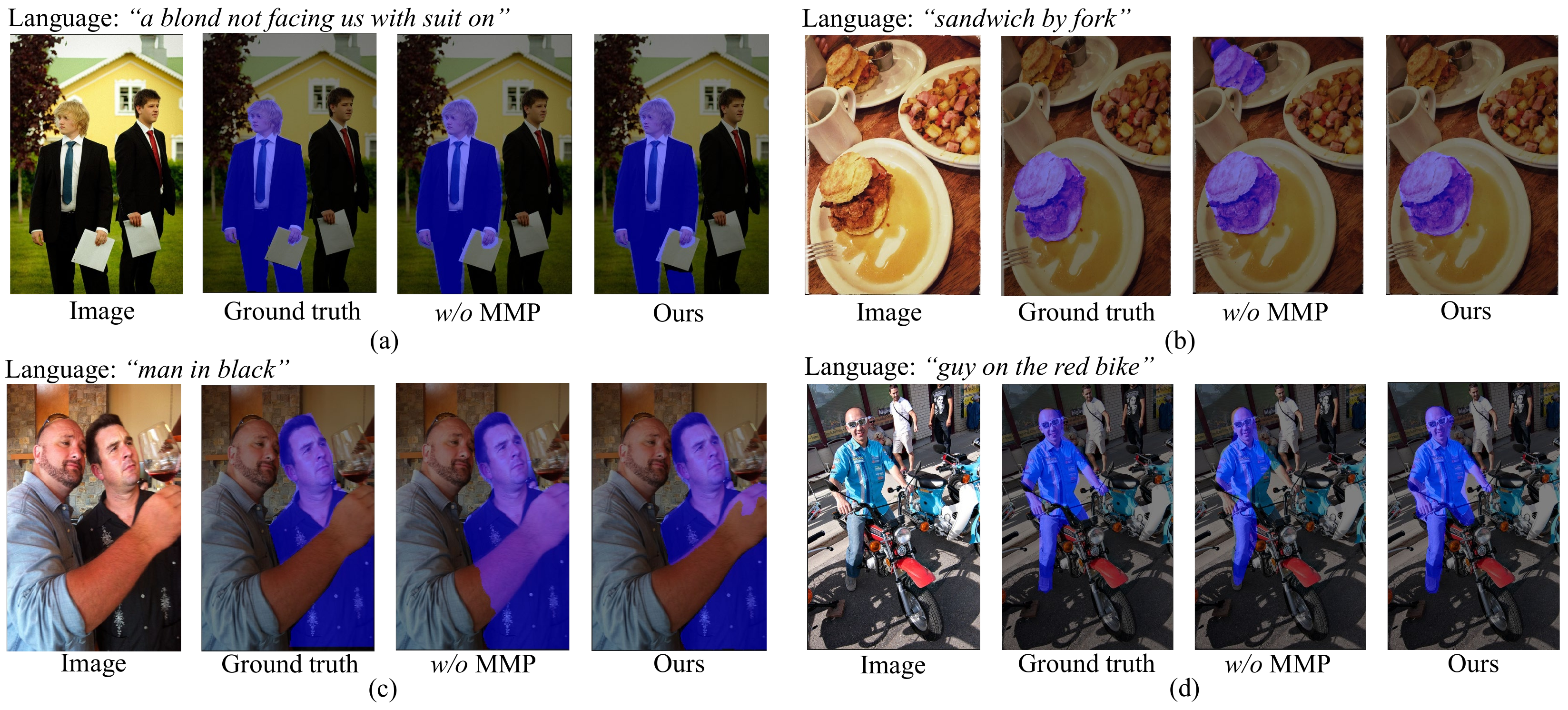}
  \caption{We generated multiple queries and use these queries to obtain corresponding segmentation mask. The final result are obtained by the weighted-sum of these masks}
\end{figure*}
\textbf{Multi-Mask Projector}. Although the benefits of using multiple queries for better performance are well known, we still want to know whether the multiple masks generated by these queries have an effect on our framework. So we conduct an experiment to cast the Multi-Mask Projector, which means we varied the query number $N_q$, but always generated a single mask. We fed the generated queries into the Multi-Query Estimator to obtain the score for each query, and used them to compute a weighted sum with the queries themselves instead of the masks generated by these queries. This approach allowed us to generate a single final query, which was aggregated from all generated queries, and used to generate a single mask. This approach is different from simply setting $N_q$ to 1 because we still generated multiple queries but only obtained one mask as the final result.  The result are reported at Table 3. According to our findings, although we generated multiple queries, the resulting performance was still insufficient because we did not generate multiple corresponding masks, which means we did not make optimal use of the information from each query. This experiment also demonstrated the effectiveness of our Multi-Mask Projector.

\textbf{Multi-Query Estimator}. We remove the Multi-Query Estimator, so the final result will be obtained by directly adding multiple masks without any scores. As shown in Table 4, removing the Multi-Query Estimator leads to a drop of 1.23 absolute points. These results demonstrate the benefit of the Multi-Query Estimator and the effectiveness of weighted sum.

\textbf{Global visual feature}. We remove the global visual feature $f_{vg}$ which means we remove the Eq 8. To be more specific, instead of fused textual features $F_{tv}$, we directly use textual features $F_t$  to generate multiple queries with dense visual features $F_{vd}$. As shown in Table 4, removing the global visual feature $f_{vg}$ leads to a drop of 1.23 absolute points which demonstrate that the global feature is a vital part of the CLIP\cite{radford2021learning} model when it comes to the referring image segmentation. 
\section{Visualization}
As shown in Figure 5, we provide visualization results with different settings to demonstrate the benefits of our Multi-Mask Projector in our proposed method. The "\emph{w/o} MMP" label denotes the absence of the Multi-Mask Projector. From the visualization results, we can observe that the absence of the Multi-Mask Projector leads to worse segmentation masks. This occurs because the baseline network fails to effectively address the randomness of the referring expressions with the corresponding regions. For instance, as illustrated in Figure 5(b), the language expression is "\emph{sandwich by fork}," but the result displays two sandwiches. In contrast, our proposed method successfully distinguishes between the sandwich with the fork and the sandwich without the fork. However, the model is still uncertain in some challenging marginal regions.
\section{Conclusion}
In this paper, we propose an end-to-end framework, Multi-Mask Network (MMNet), that effectively reduces the randomness caused by diverse objects and unrestricted language. MMNet is based on the CLIP architecture, utilizing its global and fine-grained information features. Our method generates a series of masks based on various aspects of language expression, combining them to produce the final prediction mask. This approach enhances the use of generated queries and reduces uncertainty and ambiguity of language expression. Our experiments show that MMNet significantly outperforms previous state-of-the-art methods on RefCOCO, RefCOCO+ and G-Ref datasets without any post-processing. Extensive ablation studies on three commonly used datasets have validated the effectiveness of each proposed component.
\clearpage
\printbibliography

\end{document}